\documentclass[10pt,twocolumn,letterpaper]{article}
\usepackage[margin=0.75in,columnsep=0.25in]{geometry}
\usepackage[utf8]{inputenc}
\usepackage[T1]{fontenc}
\usepackage{lmodern}
\usepackage{cite}
\usepackage{amsmath,amssymb,amsfonts}
\usepackage{graphicx}
\graphicspath{{fig/}}
\usepackage{booktabs}
\usepackage{multirow}
\usepackage{makecell}
\usepackage{needspace}
\usepackage{array}
\usepackage{url}
\usepackage[hidelinks]{hyperref}

\title{EviRef: Evidence-Specific Memories for Logical Anomaly Detection across Capture Conditions}
\author{
Yaoting Huang\textsuperscript{1}, Wenqi Cai\textsuperscript{1}, Yijin Wei\textsuperscript{1}, Jianjian Qin\textsuperscript{2}\\
Chunzhi Gu\textsuperscript{3}, Haoran Xie\textsuperscript{4}, Chao Zhang\textsuperscript{1}\\[0.5em]
\small\textsuperscript{1}University of Toyama \quad \textsuperscript{2}NingboTech University\\
\small\textsuperscript{3}University of Fukui\\
\small\textsuperscript{4}Japan Advanced Institute of Science and Technology
}
\date{}

\begin{document}
\maketitle

\begin{abstract}
Logical anomalies arise when every component looks normal but their combination violates the scene rules. In practical inspection, these rules must hold under varying backgrounds, illumination, and imaging quality, with only a few dozen normal images per scene, whereas existing methods couple normality with appearance and merge different constraints into one reference. We model logical normality as decomposable constraints and propose EviRef, a training-free method that builds a separate reference memory for each family of evidence. Component prototypes built once per scene parse images under any capture condition into one label space, on which occupancy and relation constraints are verified. Semantic constraints are verified with whole-image embeddings independent of parsing, and additional evidence such as text is admitted through a task-level gate without modifying existing memories. On VID-AD, EviRef attains a mean image-level area under the receiver operating characteristic curve (AUROC) of 92.1\% (88.9\% with visual evidence alone), the highest among the compared methods under each of five capture conditions, with a standard deviation of 0.5. Even with memories built from only 20 normal images per task, it surpasses all compared methods that use 50 normal images per task.
\end{abstract}


\section{Introduction}
\label{sec:intro}

\begin{figure}[!t]
\centering
\includegraphics[page=1,width=\columnwidth]{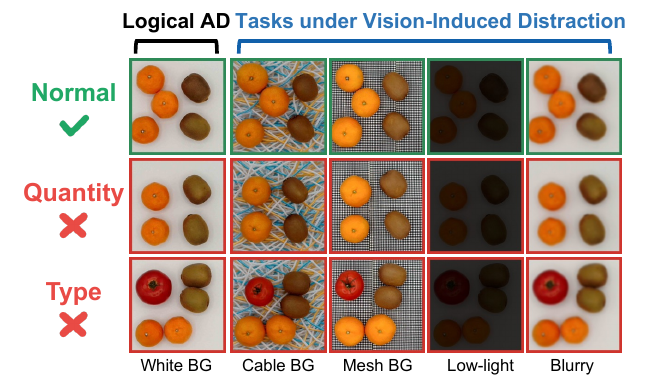}
\caption{Examples of normal images and logical anomalies under five VID-AD capture conditions. The columns show, from left to right, White BG, Cable BG, Mesh BG, Low-light, and Blurry. The rows show normal images, quantity anomalies, and type anomalies. Normal scenes contain the same fruit count and types across all five conditions.}
\label{fig:teaser}
\end{figure}

Visual anomaly detection builds a model of normality from normal samples alone, because anomalies are scarce and cannot be enumerated~\cite{pang2021survey,bergmann2019mvtec}. For structural defects such as scratches and cracks, which alter local appearance, characterizing the distribution of normal appearance suffices~\cite{roth2022patchcore,deng2022rd4ad,zavrtanik2021draem}. In assemblies, packages, and displays composed of multiple components, however, every component may look normal whereas their combination violates the rules of the scene, e.g., a wrong number, size, or category of components, or a changed position or relation~\cite{bergmann2022loco,nakata2025vidad}. The evidence of such \emph{logical anomalies}~\cite{bergmann2022loco} is not located in any local region. Instead, it resides in the constraints jointly formed by the semantic identity, the geometric occupancy, and the mutual relations of the components~\cite{liu2023comad,kim2024psad}. Detection therefore asks whether the whole image satisfies the scene rules rather than whether a local region looks normal. In deployment, the same rules must moreover hold under varying backgrounds, illumination, and imaging quality~\cite{wang2024realiad,heckler2025mvtecad2,nakata2025vidad}, whereas only a few dozen unannotated normal images are available per scene~\cite{nakata2025vidad}. Fixed rules, variable appearance, and scarce samples thus form the basic setting of logical anomaly detection in real-world scenes (Fig.~\ref{fig:teaser}).

Methods for logical anomalies have evolved in roughly three stages. Early work retained the framework of appearance-based detection, characterizing the normal distribution in feature space by reconstruction~\cite{guo2023thfr,dai2024grad}, distillation~\cite{rudolph2023ast,zhang2024cad,batzner2024efficientad}, or memory banks~\cite{cohen2023sinbad,sugawara2024puad} and capturing cross-region dependencies through a global branch~\cite{bergmann2022loco,yao2024glcf}. Component-aware methods then introduced component structure explicitly, characterizing normality by component statistics such as counts and areas~\cite{liu2023comad,kim2024psad,hsieh2024csad,peng2025samlad}. Recent vision-language methods express the scene rules through textual prompts or through descriptions, question lists, or code generated by large models~\cite{gu2024anomalygpt,zhang2024logicode,jin2025logicad,kwon2025logicqa,zhang2025logsad}. Normality is thus increasingly obtained without training, stated explicitly, and drawn from several sources of evidence.

However, these approaches still have distinct limitations. First, training-free memory-bank methods store scene rules and visual appearance in the same features, tying the reference to the capture condition. The reference can therefore shift with changes in background, illumination, or imaging quality~\cite{nakata2025vidad,heckler2025mvtecad2}, and can be difficult to estimate from scarce normal samples~\cite{huang2022regad,xie2023graphcore}. Second, component-aware methods represent statistics such as counts and areas explicitly, but often assess them with one scoring module. Since quantity, category, position, and relation constraints can be violated independently~\cite{bergmann2022loco}, combining them into one score can let the strongest cue dominate and obscure which constraint is violated. Finally, language-based methods make the rules explicit, but still require a large model to interpret and verify scene-specific prompts, questions, or code~\cite{gu2024anomalygpt,zhang2024logicode,jin2025logicad,kwon2025logicqa,zhang2025logsad}.

To address these limitations, we represent normality with explicit, decomposable constraints on scene components. These constraints specify which semantic categories are present, how much area each occupies, and how the components are arranged relative to one another. Each constraint is a group of low-dimensional statistics estimated directly from normal images and verified independently. The occupancy and relation constraints are defined on a component label space, and their descriptors are therefore independent of capture-condition appearance, whereas the semantic constraints draw on whole-image evidence that is independent of component parsing. With only a few dozen normal images per scene, EviRef estimates these low-dimensional statistics directly instead of training a discriminator~\cite{huang2022regad,xie2023graphcore,cohen2020spade,damm2025anomalydino}.

Specifically, EviRef uses open-vocabulary detection and segmentation to build component prototypes for each scene~\cite{liu2024groundingdino,kirillov2023sam}. A frozen backbone~\cite{zagoruyko2016wrn} then maps images from all capture conditions into the same component label space. EviRef builds separate memories from normal images for occupancy, semantic~\cite{radford2021clip}, and relation evidence. For each path, the discrepancy between a test image and its corresponding memory measures the violation of that constraint family. EviRef standardizes and sums these path scores to obtain an image-level anomaly score. The method can also incorporate nonvisual evidence, such as printed text on labels~\cite{baek2019craft,shi2017crnn}, through a task-level gate. This addition requires no retraining and leaves the existing memories unchanged. The contributions are as follows.
\begin{enumerate}
\item We formulate logical normality as independently verifiable occupancy, semantic, and relation constraints.
\item We develop EviRef, a training-free method that shares component prototypes across capture conditions, maintains a reference memory for each evidence family, and incorporates text through a task-level gate.
\item Across all 50 VID-AD tasks, EviRef achieves a mean image-level AUROC of 92.1\% and ranks first under all five capture conditions. With 20 normal images per task, it surpasses all compared methods that use 50 normal images per task.
\end{enumerate}

\section{Related Work}
\label{sec:related}

\subsection{Structural Anomaly Detection}
Conventional visual anomaly detection targets structural anomalies such as scratches, cracks, and contamination and models normality from normal samples alone~\cite{pang2021survey}, using memory banks or density models of pretrained features~\cite{cohen2020spade,roth2022patchcore,defard2021padim}, reconstruction or distillation~\cite{deng2022rd4ad}, anomaly synthesis~\cite{zavrtanik2021draem,liu2023simplenet}, or vision-language models~\cite{jeong2023winclip,zhou2024anomalyclip}. Few-shot variants need only a handful of them~\cite{huang2022regad,fang2023fastrecon}. These methods approach saturation on MVTec AD~\cite{bergmann2019mvtec}, BTAD~\cite{mishra2021btad}, and VisA~\cite{zou2022visa}, and Real-IAD~\cite{wang2024realiad} and MVTec AD 2~\cite{heckler2025mvtecad2} extend them to more demanding capture conditions. Their shared premise, however, is that an anomaly alters the appearance of a local region, which does not hold for logical anomalies.

\subsection{Logical Anomaly Detection}
MVTec LOCO AD~\cite{bergmann2022loco} first separated logical from structural anomalies, and VID-AD~\cite{nakata2025vidad} captures the same scenes under varying backgrounds, illumination, and imaging quality, requiring rule-based discrimination to remain effective despite visual distraction. Methods for these benchmarks fall into three groups. Appearance-based methods enlarge the receptive field through global or autoencoder branches~\cite{bergmann2022loco,batzner2024efficientad,yao2024glcf,rudolph2023ast,zhang2024cad}, use template-guided reconstruction~\cite{guo2023thfr,dai2024grad}, or estimate density on set descriptors~\cite{cohen2023sinbad,sugawara2024puad}. Component-aware methods decompose the image into components and model the normal range of component statistics with a memory bank or a scoring network~\cite{liu2023comad,kim2024psad,hsieh2024csad,peng2025samlad}. Methods based on large language or vision-language models encode the rules in textual descriptions, question lists, or generated code~\cite{gu2024anomalygpt,zhang2024logicode,jin2025logicad,kwon2025logicqa,zhang2025logsad,nakata2025vidad}.

\subsection{From Implicit Learning to Explicit Decomposition}
These method families differ in how they represent logical constraints and assess their violations. Appearance-based methods model the normal distribution as a whole in feature space. Whether implemented through training or a memory bank, these methods encode rules and appearance together in the same features, residuals, or density model, causing both to drift when the capture condition changes~\cite{nakata2025vidad,heckler2025mvtecad2}. Component-aware methods already decompose logical constraints, grounding normality in component statistics such as counts, areas, and compositions~\cite{liu2023comad,kim2024psad}. These statistics, however, are still evaluated holistically by one scoring network or memory bank~\cite{kim2024psad,hsieh2024csad} that outputs one score, which neither indicates the violated constraint nor prevents a single cue from dominating. Large-model methods write the rules out one by one as text or code, but each rule must be distilled and answered by the model for the specific scene~\cite{zhang2024logicode,kwon2025logicqa,zhang2025logsad}, and verification depends on the model's perception of fine-grained attributes such as quantity, size, and position~\cite{paiss2023count}, making stability dependent on the model itself. EviRef instead separates the constraints into occupancy, semantic, and relation families, estimates a memory for each directly from normal images, and verifies them independently without training, preserving an independent violation score for each family. TriP-AD~\cite{huang2025tripad} likewise scores occupancy, relation, and whole-image features in three paths, but treats them as complementary cues rather than as constraint families with their own memories, and uses no non-visual evidence.
\begin{figure*}[!t]
\centering
\includegraphics[page=1,width=\textwidth]{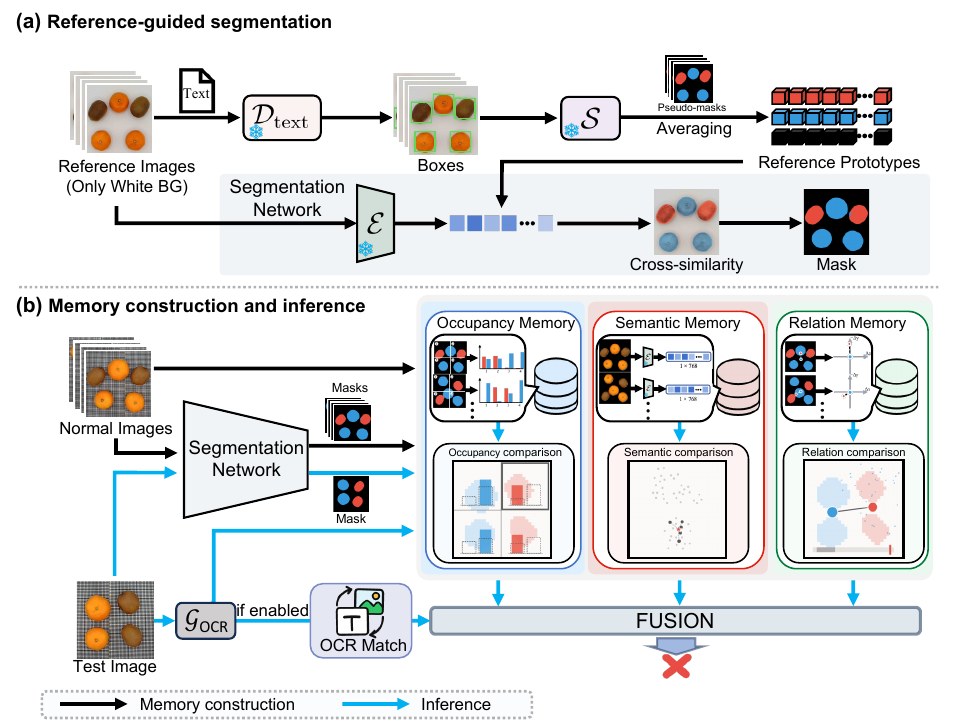}
\caption{Overview of EviRef. (a) Frozen text-guided detection and segmentation identify components in reference images from the \emph{White BG} condition. Their features form component prototypes, and nearest-prototype matching assigns images from every condition to the same component labels. (b) Normal images define separate occupancy, semantic, and relation memories. At inference, EviRef compares a test image with each memory and combines the three scores. When enabled for \emph{Ropes}, the OCR gate rank-fuses the text score with the combined visual score. Black arrows indicate memory construction, and blue arrows indicate inference.}
\label{fig:overview}
\end{figure*}

\needspace{6\baselineskip}

\section{Method}
\label{sec:method}

\subsection{Problem Setting and Overview}
\label{sec:method_paradigm}
Given a scene $s$, EviRef uses two sets of unannotated normal images. The \emph{scene-level reference set} $\mathcal{R}_s$ contains images from the \emph{White BG} condition, represented by the plain white background in Fig.~\ref{fig:teaser}. It is used once to build component prototypes shared across all capture conditions of the scene (Section~\ref{sec:method_parser}). The \emph{task-level normal set} $\mathcal{N}=\{I_1,\dots,I_N\}$ contains normal images captured under the condition being evaluated. Here, a task means a particular scene evaluated under one capture condition. This set therefore provides the normal samples used to estimate the constraint statistics for that scene and condition (Section~\ref{sec:method_paths}). The goal is to assign an image-level anomaly score $S(I)$ to a test image $I$. Scene rules remain fixed across conditions, while the background, illumination, and imaging quality vary.

Fig.~\ref{fig:overview} summarizes the EviRef pipeline. In panel (a), frozen text-guided detection and segmentation build component prototypes from normal images in the \emph{White BG} condition. Nearest-prototype matching then maps images from every capture condition into a shared component label space. In panel (b), EviRef evaluates three constraint families against separate reference memories estimated from normal images. Occupancy measures component presence and area, capturing quantity, length, and missing-component anomalies. Semantic evidence checks whether the expected component categories are present, capturing type anomalies. Relation evidence describes relative component configurations, capturing placement and relation anomalies~\cite{bergmann2022loco,nakata2025vidad}. Logical normality requires all three families to remain within their normal ranges. EviRef standardizes and combines deviations from these ranges into an image-level anomaly score $S(I)$. For \emph{Ropes}, the OCR gate rank-fuses the text score with the combined OVR score without changing the existing memories. All networks remain frozen, and the normal statistics can be estimated from a few dozen images.

\subsection{Component Parsing}
\label{sec:method_parser}
Because logical constraints are defined over scene components, EviRef first converts each image into a component map. This parsing must remain consistent across capture conditions, since condition-dependent labeling errors would affect the occupancy and relation statistics computed from the map. EviRef therefore avoids training a segmenter and instead builds component prototypes once per scene, then parses images from all conditions by nearest-prototype matching.

\textbf{Component prototypes.} The component nouns of the scene rules prompt a frozen Grounding DINO~\cite{liu2024groundingdino} to localize the components in the \emph{White-BG} reference images, and SAM-HQ~\cite{kirillov2023sam,ke2023samhq} turns the detections into instance masks. Language is used only for localization and takes no part in scoring. Since the detections are instance-level, the instances are clustered by appearance features and color into scene-level component categories, and background clusters and outliers are removed, giving a pseudo component map for every reference image. Mid-level features of a frozen WRN-50~\cite{zagoruyko2016wrn} are averaged within the mask of each category and $\ell_2$-normalized to give the prototype $\mathbf{p}_c$. The prototype set $\mathcal{P}=\{\mathbf{p}_c\}_{c=0}^{K-1}$, where class $0$ is the background, is shared by all capture conditions of the scene.

\textbf{Nearest-prototype parsing.} For any normal or test image $I$, every spatial location is assigned to the most similar prototype.
\begin{equation}
M(I)_{ij}=\arg\max_{c}\;\big\langle \hat{\mathbf{e}}_{ij}(I),\,\mathbf{p}_c\big\rangle,
\label{eq:parsing}
\end{equation}
where $\hat{\mathbf{e}}_{ij}(I)$ is the $\ell_2$-normalized feature at location $(i,j)$. Upsampled to the image resolution, this gives the component map $M(I)$. This parameter-free segmentation network (Fig.~\ref{fig:overview}) maps images under all conditions into the same label space, on which the occupancy and relation descriptors are computed.

\subsection{Constraint Memories}
\label{sec:method_paths}
Each constraint family is realized as a scoring path with its own descriptor and its own memory estimated from normal images. No statistics are shared, which is how the decomposition of constraints is realized. The path score $s_X$, $X\in\{O,V,R\}$, is the degree of violation of the family. O and R are computed on the component map, whereas V uses whole-image embeddings and complements them with evidence that is independent of parsing.

\textbf{Occupancy path (O).} On $M(I)$, the pixel proportion of each component category at two scales $g$, the whole image and a $2\times2$ partition, gives the occupancy histograms $\mathbf{h}^{(g)}(I)$. With the mean $\boldsymbol{\mu}_O^{(g)}$ and covariance $\boldsymbol{\Sigma}_O^{(g)}$ estimated from normal images,
\begin{equation}
s_O(I)=\sum_{g} d_{\mathrm{M}}\big(\mathbf{h}^{(g)}(I),\,\boldsymbol{\mu}_O^{(g)},\boldsymbol{\Sigma}_O^{(g)}\big),
\label{eq:occupancy}
\end{equation}
where $d_{\mathrm{M}}$ is the Mahalanobis distance, computed with the Moore-Penrose pseudo-inverse since the entries of a histogram sum to one. Adding, removing, or lengthening a component directly changes the histograms. The whole-image histogram reflects the total amount of each category, and the $2\times2$ partition also reflects where the amount changes.

\textbf{Semantic path (V).} A replacement category absent from the normal images has no prototype and may be parsed as an existing category, leaving the component map unchanged, preventing either O or R from responding. The semantic path therefore takes its evidence from the whole image. The frozen CLIP image encoder~\cite{radford2021clip} (ViT-L/14 at a resolution of 336) gives an embedding $\mathbf{v}(I)$, the embeddings $\{\mathbf{v}(I_n)\}$ of the normal images form the memory, and
\begin{equation}
s_V(I)=\frac{1}{k}\sum_{n\in\mathcal{N}_k(I)}\big\|\mathbf{v}(I)-\mathbf{v}(I_n)\big\|_2,\qquad k=5,
\label{eq:semantic}
\end{equation}
where $\mathcal{N}_k(I)$ indexes the $k$ nearest normal embeddings. Encoding the whole image rather than component crops keeps the path free of parsing errors.

\textbf{Relation path (R).} For every pair of component categories $(a,b)$, a geometric descriptor is computed from the centroids and areas of the two regions,
\begin{equation}
\mathbf{r}_{ab}(I)=\big[\Delta x,\;\Delta y,\;\sqrt{\Delta x^2+\Delta y^2},\;A_a/A_b\big]^{\top},
\label{eq:relation_desc}
\end{equation}
with the displacements normalized by the image size and $A$ the area. Pairs that co-occur in fewer than 10\% of the normal images are discarded. For each kept pair a Gaussian $(\boldsymbol{\mu}_{ab},\boldsymbol{\Sigma}_{ab})$ is estimated from normal images, and a missing member of an expected pair contributes a fixed distance of 2.0 to the maximum in (\ref{eq:relation}). The path score is
\begin{equation}
s_R(I)=\max_{(a,b)}\; d_{\mathrm{M}}\big(\mathbf{r}_{ab}(I),\boldsymbol{\mu}_{ab},\boldsymbol{\Sigma}_{ab}\big).
\label{eq:relation}
\end{equation}
The maximum is used because a relation violation usually involves a single pair, which a sum would dilute. If a scene has no stably co-occurring pair, $s_R$ is constant. Its standardized score is therefore set to zero, removing this path from the fusion.

\subsection{Score Standardization and Integration of Additional Evidence}
\label{sec:method_fusion}
This subsection explains how EviRef combines heterogeneous degrees of violation without learnable weights and admits new sources of evidence without modifying existing memories. New evidence may be \emph{memory-based}, i.e., like O, V, and R, it estimates a normal range from normal images, or \emph{self-consistency-based}, i.e., it checks within each image whether different sources of information agree and needs no normal statistics.

\textbf{Standardization on normal statistics.} The path scores have different units. For each path, a trimmed mean $\tilde{\mu}_X$ and standard deviation $\tilde{\sigma}_X$ are estimated from the scores of the normal images between the 20th and 80th percentiles, and
\begin{equation}
\hat{s}_X(I)=\frac{s_X(I)-\tilde{\mu}_X}{\tilde{\sigma}_X},\qquad S_{\text{OVR}}(I)=\sum_{X\in\{O,V,R\}} \hat{s}_X(I).
\label{eq:fusion}
\end{equation}
Trimming keeps the estimate unaffected by the few normal images whose parsing fails. Each standardized score expresses the deviation in multiples of the path's own normal fluctuation. Accordingly, $S_{\text{OVR}}$ is an equal-weight sum without learnable weights, and a new memory-based path only needs its own normal statistics.

\textbf{Gating textual evidence.} Some scene rules depend on text. For example, in the \emph{Ropes} scene of VID-AD, the color of the rope must match the color word printed on the label. Such evidence is admitted through a task-level gate determined only from the normal images of the task. EasyOCR, which combines a CRAFT detector~\cite{baek2019craft} and a CRNN recognizer~\cite{shi2017crnn}, measures the proportion of images with reliable text and the proportion whose text contains a color word. The gate opens when these proportions are at least $0.3$ and $0.1$, respectively. The gate is fixed before any test image is seen, and the second condition excludes OCR false detections on blurred textures.

\textbf{Matching text and image content.} Once the gate is open, the color word $\ell$ of each test image is recognized, the most elongated component category in $M(I)$ is taken as the rope region, and its CLIP image embedding $\mathbf{u}$ is compared with the CLIP text embeddings $\mathbf{t}_c$ of the colors.
\begin{equation}
s_{\text{OCR}}(I)=\max_c \operatorname{sim}(\mathbf{u},\mathbf{t}_c)-\operatorname{sim}(\mathbf{u},\mathbf{t}_\ell).
\label{eq:ocr}
\end{equation}
A positive value indicates that the text contradicts the object. If no color word can be read, $s_{\text{OCR}}$ receives its maximal value, consistent with the missing-pair penalty of R.

\textbf{Rank-based fusion.} Unlike the visual paths, $s_{\text{OCR}}$ is self-consistency-based evidence. It measures disagreement between the recognized color word and the rope's visual appearance within each image and requires no reference memory. Because its scale is not calibrated against a normal-image score distribution, we combine it with $S_{\text{OVR}}$ by ranking both scores.
\begin{equation}
S(I)=\begin{cases}
S_{\text{OVR}}(I), & \text{gate off},\\[3pt]
\begin{aligned}
&\tfrac{1}{2}\big[\operatorname{rank}\big(S_{\text{OVR}}(I)\big)\\
&\quad+\operatorname{rank}\big(s_{\text{OCR}}(I)\big)\big],
\end{aligned} & \text{gate on},
\end{cases}
\label{eq:final}
\end{equation}
where $\operatorname{rank}(\cdot)$ is the rank among the test images of the task, normalized to $[0,1]$ with ties averaged, making it a monotone transform within the task that uses no labels. Because the ranks are taken within the test images of a task, the gate-on score assumes offline inspection of an unlabeled batch. Tasks with the gate off are scored image by image. All three visual paths remain in $S_{\text{OVR}}$, and none of their statistics is changed.

\section{Experiments}
\label{sec:exp}

\subsection{Experimental Setup}
\label{sec:exp_setup}
\textbf{Datasets and metric.} VID-AD~\cite{nakata2025vidad} contains 10 scenes, each captured under five conditions. These comprise a white background (\emph{White BG}), a cable background (\emph{Cable BG}), a mesh background (\emph{Mesh BG}), low light (\emph{Low-light CD}), and blur (\emph{Blurry CD}), for 50 scene-condition combinations. Each task provides 50 normal training images, and its test set contains 50 normal and 100 to 111 logically anomalous images. Across the dataset, anomalies fall into six types, namely quantity, length, type, placement, relation, and mix. Mix cases contain several co-occurring anomalies. Each scene involves two basic violation types and their combination. The scene rules remain unchanged across the five conditions and only the background, illumination, and imaging quality vary, which matches the setting of fixed rules and variable appearance considered in this paper. We also evaluate on the logical anomalies of MVTec LOCO AD~\cite{bergmann2022loco}. We report the image-level AUROC. For VID-AD, we average AUROC for each condition over the 10 scenes, then compute Mean and Std across the five conditions. Std measures variation across capture conditions.

\textbf{Compared methods.} On VID-AD, the appearance-based methods are VAE~\cite{kingma2014vae}, AnoGAN~\cite{schlegl2017anogan}, PaDiM~\cite{defard2021padim}, SPADE~\cite{cohen2020spade}, PatchCore~\cite{roth2022patchcore}, and EfficientAD~\cite{batzner2024efficientad}. The component-aware methods are ComAD~\cite{liu2023comad} and CSAD~\cite{hsieh2024csad}. We also compare with the set-feature method SINBAD~\cite{cohen2023sinbad}, the foundation-model-based methods LogSAD~\cite{zhang2025logsad} and UniVAD~\cite{gu2025univad}, the method of Nakata \emph{et al.}~\cite{nakata2025vidad}, and TriP-AD~\cite{huang2025tripad}. We quote the results for the last two from their papers. We run all other methods with their official implementations on the same images. For LogSAD, we omit its hand-written dataset-specific rules because they do not apply to VID-AD. We use UniVAD in its default one-shot setting.

\textbf{Implementation details.} Images are resized to $256\times256$. Component parsing uses the concatenated layer-2 and upsampled layer-3 features of WRN-50 on a $32\times32$ grid. The confidence threshold for reliable text in the OCR gate is $0.3$. All other hyperparameters are given in Section~\ref{sec:method} and fixed for all scenes and conditions. The memories and standardization statistics are estimated from the normal training images of the current task.

\begin{figure*}[!t]
\centering
\includegraphics[page=1,width=\textwidth]{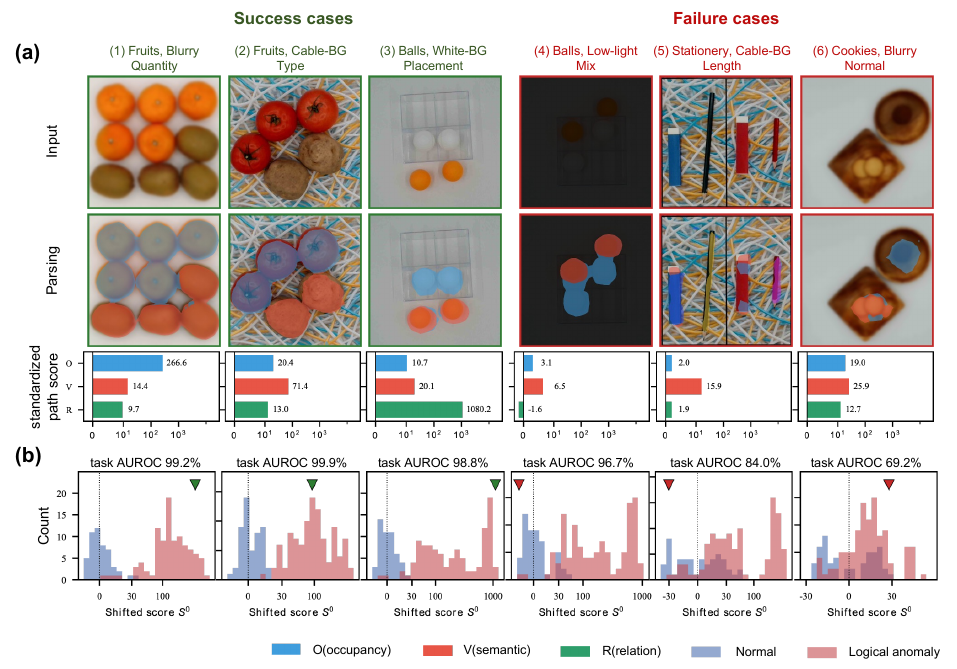}
\caption{Qualitative results on VID-AD. (a) The first three examples are correctly detected anomalies. The last three are errors, consisting of two missed anomalies and one normal image incorrectly flagged as anomalous. Each example is shown with its input image, component parsing, and standardized scores $\hat{s}_X$ for the occupancy (O), semantic (V), and relation (R) paths. The headings identify each example's scene, capture condition, and anomaly type. Parsing colors denote scene-specific component categories. For visualization, feature-map similarities are upsampled to image resolution before each pixel is assigned to its nearest component prototype. (b) Score distributions of normal and anomalous test images for the six corresponding tasks. Each triangle marks the score of the example shown above. Scores are shifted by subtracting the median normal-image score for the corresponding task. This shift preserves image ranking and therefore does not change AUROC.}
\label{fig:qual}
\end{figure*}

\subsection{Comparison with the State of the Art}
\label{sec:exp_main}
\begin{table}[!t]
\centering
\caption{\upshape Image-level AUROC (\%) of anomaly-detection methods on VID-AD. Values are averaged over 10 scenes. White, Cable, and Mesh denote White BG, Cable BG, and Mesh BG. Low and Blur denote Low-light CD and Blurry CD. Mean and Std summarize performance across the five conditions. Bold and underlined values mark the best and second-best results.}
\label{tab:main}
\setlength{\tabcolsep}{2pt}\fontsize{8.5}{10}\selectfont
\resizebox{\columnwidth}{!}{
\begin{tabular}{@{}l ccccc cc@{}}
\toprule
Method & \emph{White} & \emph{Cable} & \emph{Mesh} & \emph{Low} & \emph{Blur} & Mean$\uparrow$ & Std$\downarrow$ \\
\midrule
VAE~\cite{kingma2014vae} & 49.3 & 59.2 & 45.5 & 48.6 & 47.2 & 49.9 & 4.8 \\
AnoGAN~\cite{schlegl2017anogan} & 49.3 & 53.1 & 54.1 & 45.4 & 49.3 & 50.3 & 3.1 \\
PaDiM~\cite{defard2021padim} & 34.8 & 35.4 & 42.1 & 35.4 & 40.3 & 37.6 & 3.0 \\
SPADE~\cite{cohen2020spade} & 64.1 & 61.4 & 68.9 & 63.7 & 64.9 & 64.6 & 2.5 \\
PatchCore~\cite{roth2022patchcore} & 49.5 & 50.3 & 49.9 & 49.0 & 52.8 & 50.3 & \underline{1.3} \\
EfficientAD~\cite{batzner2024efficientad} & 57.1 & 51.5 & 58.8 & 58.8 & 59.9 & 57.2 & 3.0 \\
ComAD~\cite{liu2023comad} & 79.3 & 78.5 & 76.8 & 76.7 & 81.0 & 78.5 & 1.6 \\
ComAD+PatchCore~\cite{liu2023comad} & 65.2 & 58.6 & 64.1 & 62.2 & 65.6 & 63.2 & 2.5 \\
SINBAD~\cite{cohen2023sinbad} & 81.0 & 82.9 & 80.6 & 79.4 & 84.9 & 81.8 & 1.9 \\
CSAD~\cite{hsieh2024csad} & 87.8 & 76.1 & 83.5 & 81.0 & 81.0 & 81.9 & 3.8 \\
LogSAD~\cite{zhang2025logsad} & 69.4 & 63.9 & 70.9 & 69.4 & 69.4 & 68.6 & 2.4 \\
UniVAD~\cite{gu2025univad} & 61.6 & 62.1 & 61.4 & 62.4 & 56.3 & 60.7 & 2.3 \\
Nakata \emph{et al.}~\cite{nakata2025vidad} & 82.5 & 81.1 & \underline{84.8} & 84.2 & 82.6 & 83.1 & \underline{1.3} \\
TriP-AD~\cite{huang2025tripad}& \underline{88.4} & \underline{85.1} & \underline{84.8} & \underline{88.4} & \underline{87.4} & \underline{86.8} & 1.6 \\
\midrule
\textbf{EviRef (Ours)} & \textbf{93.0} & \textbf{91.6} & \textbf{91.8} & \textbf{92.3} & \textbf{91.7} & \textbf{92.1} & \textbf{0.5} \\
\bottomrule
\end{tabular}
}
\end{table}

\begin{table}[!t]
\centering
\caption{\upshape Image-level AUROC (\%) of EviRef for each VID-AD scene and capture condition. Mean is the average across the five conditions. An asterisk marks the Ropes tasks where the OCR gate is enabled.}
\label{tab:per_scenario}
\setlength{\tabcolsep}{4.5pt}\fontsize{8.5}{10}\selectfont
\resizebox{\columnwidth}{!}{
\begin{tabular}{@{}l*{6}{c}@{}}
\toprule
Scene & \emph{White} & \emph{Cable} & \emph{Mesh} & \emph{Low} & \emph{Blur} & Mean \\
\midrule
\emph{Sticks} & 99.6 & 97.8 & 98.0 & 98.0 & 98.1 & 98.3 \\
\emph{Fruits} & 99.9 & 99.9 & 99.7 & 99.3 & 99.2 & 99.6 \\
\emph{Tools} & 99.7 & 98.6 & 97.1 & 99.7 & 98.5 & 98.7 \\
\emph{Cookies} & 72.6 & 70.9 & 73.2 & 71.8 & 69.2 & 71.6 \\
\emph{Tapes} & 99.0 & 100.0 & 94.0 & 98.1 & 98.8 & 98.0 \\
\emph{Stationery} & 98.8 & 84.0 & 91.3 & 98.7 & 96.4 & 93.8 \\
\emph{Ropes}$^\ast$ & 74.1 & 70.8 & 76.2 & 74.5 & 68.8 & 72.9 \\
\emph{Blocks} & 90.1 & 98.5 & 93.1 & 87.5 & 93.0 & 92.4 \\
\emph{Dishes} & 97.5 & 99.6 & 97.5 & 98.5 & 96.8 & 98.0 \\
\emph{Balls} & 98.8 & 96.4 & 98.5 & 96.7 & 98.3 & 97.7 \\
\midrule
Mean & 93.0 & 91.6 & 91.8 & 92.3 & 91.7 & 92.1 \\
\bottomrule
\end{tabular}
}
\end{table}

Table~\ref{tab:main} reports the results on VID-AD. EviRef achieves a mean AUROC of 92.1\% and is the best of the compared methods, all of which learn from unannotated normal images, under all five conditions. It improves on the method of Nakata \emph{et al.}\ (83.1\%) by 9.0 points and on the strongest compared method (86.8\%) by 5.3 points, of which 2.1 points come from the three visual paths and 3.2 points from the textual evidence on the five \emph{Ropes} tasks (Section~\ref{sec:exp_ablation}). Appearance-based methods stay at or below about 65\% because every component of a logical anomaly looks normal, methods that model components or sets of features reach up to about 82\%, and the method of Nakata \emph{et al.}\ and TriP-AD reach 83.1\% and 86.8\% with standard deviations of 1.3 and 1.6 across conditions. EviRef raises the mean while reducing the fluctuation across conditions to 0.5, the lowest among all methods. The contrast is clearest on \emph{Cable BG}, where CSAD, which trains a component segmentation network, falls from 87.8\% under \emph{White BG} to 76.1\%, whereas EviRef falls only from 93.0\% to 91.6\%. Once the occupancy and relation constraints are built on the shared component label space and the semantic constraints are verified against the memory of each condition, changes of the capture condition hardly affect the decision.

Table~\ref{tab:per_scenario} gives the results per scene. Eight of the ten scenes reach a mean AUROC of at least 92.4\%, and six exceed 97\%. The remaining errors are concentrated in \emph{Cookies} (71.6\%) and \emph{Ropes} (72.9\%), whose scores stay low under all five conditions (69.2\% to 73.2\% and 68.8\% to 76.2\%), indicating that their difficulty lies in the scenes rather than in the capture conditions. \emph{Ropes} is also the only scene whose rule involves printed text, and it remains among the hardest even with the textual evidence. The largest condition-dependent drop occurs for \emph{Stationery} under \emph{Cable BG} (84.0\%, compared with 98.8\% under \emph{White BG}). No other scene-condition pair drops by more than 7.5 points from its \emph{White BG} result.

\textbf{Qualitative results.} Fig.~\ref{fig:qual} shows three successful and three failed cases, together with the score distribution of the task to which each case belongs. Panel (a) shows how each path responds to the case, and panel (b) places the case within the score distribution of its task, which relates a single image to the task-level AUROC. In the successful cases, the path with the strongest response coincides with the violated constraint. (1) Extra fruits under \emph{Blurry CD} raise the occupancy score ($\hat{s}_O=266.6$). (2) On \emph{Cable BG}, the \emph{tomatoes} and the \emph{potato} that replace normal fruits are parsed into the existing \emph{kiwi} and \emph{orange} labels, causing only moderate changes in occupancy and relation, while the whole-image semantic path still detects the anomaly ($\hat{s}_V=71.4$). (3) Two orange balls outside the box raise the relation score ($\hat{s}_R=1080.2$). In each case, the other two paths also respond, though more weakly ($\hat{s}_X$ between 9.7 and 20.4), in line with the complementary contributions analyzed in Section~\ref{sec:exp_type}. All three images score higher than every normal test image of their tasks.

The failed cases fall into three categories. (4) Under \emph{Low-light CD}, the parser merges adjacent balls into one connected region. Consequently, the moved ball no longer changes the relation descriptor ($\hat{s}_R=-1.6$), and the image scores lower than every normal image of its task. Since the task still reaches an AUROC of 96.7\%, this is an isolated parsing error rather than a failure of the constraint itself. (5) On \emph{Cable BG}, the change in component length is too subtle to be reflected in either the occupancy grid or the area ratio. Only the semantic path responds weakly ($\hat{s}_V=15.9$), and the image scores higher than only 12\% of the normal images. Such attributes lie beyond what the current descriptors can express. (6) A normal \emph{Cookies} image with the tray rotated by about 45$^\circ$ raises all three path scores and scores higher than 86\% of the anomalous images of its task. Because the occupancy and relation constraints are defined on a fixed spatial partition and image coordinates, the rotation is treated as a violation. This is the viewpoint assumption discussed in Section~\ref{sec:conclusion}, and it is consistent with the low AUROC of this task (69.2\%). The three failures thus stem from distinct causes. They reflect a parsing error, limited descriptor expressiveness, and the fixed-viewpoint assumption.

\subsection{Ablation Study}
\label{sec:exp_ablation}
\begin{table}[!t]
\centering
\caption{\upshape Image AUROC (\%) for EviRef with different combinations of occupancy (O), semantic (V), relation (R), and OCR evidence on VID-AD. The OCR gate is enabled only for the five Ropes tasks.}
\label{tab:abl_component}
\setlength{\tabcolsep}{2.2pt}\fontsize{8.5}{10}\selectfont
\resizebox{\columnwidth}{!}{
\begin{tabular}{@{}cccc ccccc cc@{}}
\toprule
O & V & R & Gate & \emph{White} & \emph{Cable} & \emph{Mesh} & \emph{Low} & \emph{Blur} & Mean$\uparrow$ & Std$\downarrow$ \\
\midrule
$\checkmark$ & & & & 83.4 & 81.1 & 81.0 & 82.8 & 80.7 & 81.8 & 1.1 \\
 & $\checkmark$ & & & 69.4 & 76.9 & 72.1 & 69.1 & 69.9 & 71.5 & 2.9 \\
 & & $\checkmark$ & & 81.5 & 82.5 & 78.8 & 81.3 & 82.1 & 81.2 & 1.3 \\
 & $\checkmark$ & $\checkmark$ & & 88.9 & \underline{87.8} & 86.7 & 88.9 & 88.6 & 88.2 & \underline{0.9} \\
$\checkmark$ & & $\checkmark$ & & 86.4 & 84.8 & 83.6 & 86.5 & 85.8 & 85.4 & 1.1 \\
$\checkmark$ & $\checkmark$ & & & 87.6 & 84.2 & 86.4 & 85.2 & 83.6 & 85.4 & 1.5 \\
\midrule
$\checkmark$ & $\checkmark$ & $\checkmark$ & & \underline{90.0} & 87.4 & \underline{88.0} & \underline{90.1} & \underline{88.8} & \underline{88.9} & 1.1 \\
$\checkmark$ & $\checkmark$ & $\checkmark$ & $\checkmark$ & \textbf{93.0} & \textbf{91.6} & \textbf{91.8} & \textbf{92.3} & \textbf{91.7} & \textbf{92.1} & \textbf{0.5} \\
\bottomrule
\end{tabular}
}
\end{table}

\begin{table}[!t]
\centering
\caption{\upshape Image AUROC (\%) comparing per-condition memories (PC) with memories built once from White BG images (WB). Mean (unseen) is averaged over the four conditions other than White BG. $\Delta$ is the change from PC to WB. Small White BG differences (at most 0.3 points) reflect run-to-run variation in R. Bold marks the best result in each row. Underlining marks the second-best Std or $\Delta$. Lower Std and a smaller absolute $\Delta$ are better.}
\label{tab:abl_memory}
\setlength{\tabcolsep}{1.2pt}\fontsize{8.5}{10}\selectfont
\resizebox{\columnwidth}{!}{
\begin{tabular}{@{}l cc cc cc cc cc@{}}
\toprule
 & \multicolumn{2}{c}{O only} & \multicolumn{2}{c}{V only} & \multicolumn{2}{c}{R only} & \multicolumn{2}{c}{O+V+R} & \multicolumn{2}{c}{EviRef} \\
\cmidrule(lr){2-3}\cmidrule(lr){4-5}\cmidrule(lr){6-7}\cmidrule(lr){8-9}\cmidrule(l){10-11}
Condition & PC & WB & PC & WB & PC & WB & PC & WB & PC & WB \\
\midrule
White & 83.4 & 83.4 & 69.4 & 69.4 & 81.5 & 81.7 & 90.0 & 89.7 & \underline{93.0} & \textbf{93.1} \\
Cable & 81.1 & 77.1 & 76.9 & 58.5 & 82.5 & 78.5 & \underline{87.4} & 81.7 & \textbf{91.6} & 83.0 \\
Mesh  & 81.0 & 74.0 & 72.1 & 58.8 & 78.8 & 76.3 & \underline{88.0} & 81.8 & \textbf{91.8} & 85.2 \\
Low   & 82.8 & 82.3 & 69.1 & 58.5 & 81.3 & 79.8 & \underline{90.1} & 84.6 & \textbf{92.3} & 88.1 \\
Blur  & 80.7 & 78.7 & 69.9 & 66.7 & 82.1 & 80.7 & \underline{88.8} & 83.7 & \textbf{91.7} & 87.7 \\
\midrule
Mean  & 81.8 & 79.1 & 71.5 & 62.4 & 81.2 & 79.4 & \underline{88.9} & 84.3 & \textbf{92.1} & 87.4 \\
Std   & \underline{1.1} & 3.4 & 2.9 & 4.7 & 1.3 & 1.9 & \underline{1.1} & 2.9 & \textbf{0.5} & 3.4 \\
\makecell[l]{Mean\\(unseen)} & 81.4 & 78.0 & 72.0 & 60.6 & 81.2 & 78.8 & \underline{88.6} & 83.0 & \textbf{91.9} & 86.0 \\
$\Delta$ & N/A & \underline{$-3.4$} & N/A & $-11.4$ & N/A & \textbf{$-2.4$} & N/A & $-5.6$ & N/A & $-5.9$ \\
\bottomrule
\end{tabular}
}
\end{table}

\textbf{Every constraint family contributes.} In Table~\ref{tab:abl_component}, the occupancy, relation, and semantic paths alone reach 81.8\%, 81.2\%, and 71.5\%. Every two-path combination (85.4\% to 88.2\%) outperforms each of its single paths, and all three reach 88.9\%, higher than any combination of two. The semantic path is the weakest alone but the most complementary. Adding V raises R from 81.2\% to 88.2\%, the best two-path result, and O from 81.8\% to 85.4\%. This is because V responds to type anomalies that leave the component map unchanged (Section~\ref{sec:method_paths}). Each constraint family thus captures anomalies that the others miss, demonstrating that the decomposition is effective rather than redundant. Rank-fusing the OCR score raises the mean to 92.1\% and reduces the standard deviation from 1.1 to 0.5. The gate opens only on the five \emph{Ropes} tasks. The gain therefore comes entirely from \emph{Ropes}, and the other 45 tasks remain unchanged. Without textual evidence, the visual paths (88.9\%) still exceed every compared method in Table~\ref{tab:main}.

\textbf{Robustness stems from the explicit constraints.} Table~\ref{tab:abl_memory} builds the memories either from the normal set $\mathcal{N}$ of each condition or once from the \emph{White-BG} reference set ($\mathcal{N}=\mathcal{R}_s$) for all five conditions. Since the latter memories are built from \emph{White-BG} images, we evaluate the transfer on the four remaining conditions (Mean (unseen)). There, EviRef attains 86.0\% without any normal image of the target condition, higher than the method of Nakata \emph{et al.}\ (83.2\%) under every one of the four conditions and close to TriP-AD (86.4\%), both of which use 50 normal images of each condition. Compared with per-condition memories (91.9\%), the transfer costs 5.9 points. The loss is largest on the textured \emph{Cable BG} and \emph{Mesh BG} (83.0\% and 85.2\%) and smaller under \emph{Low-light CD} and \emph{Blurry CD} (88.1\% and 87.7\%), indicating that background changes affect the transfer more than changes in illumination or sharpness. The drop is concentrated on the semantic path V ($-11.4$), whereas O and R drop by only 3.4 and 2.4 points, respectively. Their descriptors are defined on the component label space and contain no pixel appearance, while V is the only path that directly uses appearance. This agrees with the analysis in Section~\ref{sec:method}, which attributes EviRef's robustness mainly to explicit constraints on the component label space rather than to a particular representation.

\subsection{Constraint Decomposition and Anomaly Types}
\label{sec:exp_type}
\begin{figure*}[!t]
\centering
\includegraphics[page=1,width=\textwidth]{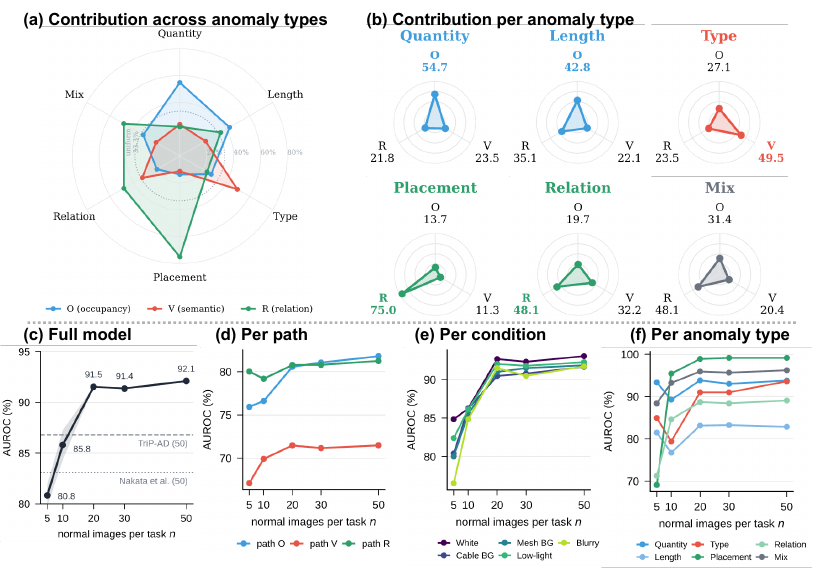}
\caption{Constraint decomposition and sample efficiency on VID-AD. For each anomalous image, negative standardized path scores are set to zero, and the remaining scores are normalized to sum to 100\%. Panel (a) overlays the mean contributions of the occupancy (O), semantic (V), and relation (R) paths for all six anomaly types. Panel (b) shows a separate radar plot for each anomaly type. Panels (c) through (f) show image AUROC versus the number $n$ of normal images per task. Panel (c) compares EviRef with TriP-AD and Nakata \emph{et al.}, both reported using 50 normal images per task. Panels (d), (e), and (f) show results for each path, capture condition, and anomaly type, respectively.}
\label{fig:analysis}
\end{figure*}
This subsection examines whether the decomposition of constraints corresponds to the actual structure of the anomalies. For every anomalous image, the negative standardized path scores are clipped to zero and normalized to sum to 100\%, and averaging over the images of a type gives the contribution of each path. Under equal contribution, each path would receive 33.3\%. Fig.~\ref{fig:analysis}(a) plots the six anomaly types in one coordinate system. The profiles of the three paths peak at different types, with O peaking at quantity, V at type, and R at placement and relation anomalies. Every basic type therefore has a dominant path, and this path coincides with the constraint that the type violates by definition.

The values in Fig.~\ref{fig:analysis}(b) show that this correspondence is not specified by hand but emerges from the data. Quantity anomalies receive their main contribution from O (54.7\%), with V and R contributing only 23.5\% and 21.8\%, because adding or removing a component directly changes the occupancy histogram. A type replacement is not always reflected in the component map, since the substituted component may be parsed as the category it replaces. Type anomalies therefore rely mainly on V (49.5\%), which takes its evidence directly from the whole image. Placement anomalies rely most heavily on R (75.0\%), with O and V contributing only 13.7\% and 11.3\%. For the remaining three types, a second path contributes more than 30\%. Relation anomalies are dominated by R (48.1\%), but V accounts for 32.2\%, since a change in the relations between components often changes the semantics of the scene. Length anomalies are dominated by O (42.8\%), with R accounting for 35.1\%, because a lengthened component changes both the occupancy histogram and the area ratio in the relation descriptor. Mix anomalies are shared by R (48.1\%) and O (31.4\%), consistent with their containing several violations. In these cases, the paths verify the same anomaly from different aspects and corroborate each other. The decision therefore combines complementary evidence, indicating that each path captures a distinct facet of the anomaly. Taking the path with the largest standardized score as the violated constraint reported by EviRef, it coincides with the constraint violated by definition for 72.7\% of the 4420 non-mix anomalous images of the 45 gate-off tasks, far above the chance level of 33.3\%.

\subsection{Sample Efficiency with Few Normal Images}
\label{sec:exp_fewnormal}
\begin{table}[!t]
\centering
\caption{\upshape Image AUROC (\%) versus the number $n$ of normal images per VID-AD task used to build the memories and standardization statistics. For $n\le10$, the Mean column reports mean $\pm$ standard deviation over three subsets. Prototypes and hyperparameters are fixed to the EviRef setting with $n=50$.}
\label{tab:fewnormal}
\setlength{\tabcolsep}{2.5pt}\fontsize{8.5}{10}\selectfont
\resizebox{\columnwidth}{!}{
\begin{tabular}{@{}c ccccc cc@{}}
\toprule
$n$ & \emph{White} & \emph{Cable} & \emph{Mesh} & \emph{Low} & \emph{Blur} & Mean & Std \\
\midrule
5 & 84.8 & 80.4 & 80.0 & 82.4 & 76.5 & 80.8 $\pm$ 1.1 & 2.8 \\
10 & 86.2 & 86.2 & 85.5 & 86.1 & 84.8 & 85.8 $\pm$ 1.5 & 0.5 \\
20 & 92.7 & 90.5 & 91.0 & 92.0 & 91.5 & 91.5 & 0.8 \\
30 & 92.3 & 90.8 & 91.5 & 91.8 & 90.5 & 91.4 & 0.7 \\
50 (EviRef) & \textbf{93.0} & \textbf{91.6} & \textbf{91.8} & \textbf{92.3} & \textbf{91.7} & \textbf{92.1} & \textbf{0.5} \\
\bottomrule
\end{tabular}
}
\end{table}

\begin{table}[!htb]
\centering
\caption{\upshape Image-level AUROC (\%) comparing training-free methods on logical anomalies in MVTec LOCO AD. The Shots column gives the number of normal reference images. SPADE and PatchCore results are from~\cite{cohen2023sinbad}, and other results are from the original papers. Abbreviations are B. Box for Breakfast Box, J. Bot. for Juice Bottle, Push. for Pushpins, S. Bag for Screw Bag, and S. Conn. for Splicing Connectors.}
\label{tab:loco_logical}
\setlength{\tabcolsep}{0.9pt}\fontsize{8.5}{10}\selectfont
\resizebox{\columnwidth}{!}{
\begin{tabular}{l c ccccc c}
\toprule
Method & Shots & \emph{B.\ Box} & \emph{J.\ Bot.} & \emph{Push.} & \emph{S.\ Bag} & \emph{S.\ Conn.} & Mean \\
\midrule
SPADE~\cite{cohen2020spade} & full & 81.8 & 91.9 & 60.5 & 46.8 & 73.8 & 71.0 \\
PatchCore~\cite{roth2022patchcore} & full & 77.7 & 83.7 & 62.2 & 55.3 & 63.3 & 68.4 \\
ComAD~\cite{liu2023comad} & full & 94.7 & 90.9 & 89.0 & 79.7 & 84.4 & 87.7 \\
ComAD+PatchCore~\cite{liu2023comad} & full & 91.1 & 95.0 & 95.7 & 71.9 & \textbf{93.3} & 89.4 \\
SINBAD~\cite{cohen2023sinbad} & full & \textbf{97.7} & \underline{97.1} & 88.9 & 81.1 & 91.5 & \underline{91.2} \\
LogSAD~\cite{zhang2025logsad} & full & N/A & N/A & N/A & N/A & N/A & 89.3 \\
SAM-LAD~\cite{peng2025samlad} & full & 96.7 & \textbf{98.7} & 97.2 & \textbf{95.2} & 91.4 & \textbf{95.8} \\
\midrule
LogicAD~\cite{jin2025logicad} & 1 & 93.1 & 81.6 & \underline{98.1} & 83.8 & 73.4 & 86.0 \\
LogicQA~\cite{kwon2025logicqa} & 3 & 87.6 & 88.2 & \textbf{98.4} & 71.5 & \underline{92.4} & 87.6 \\
\midrule
\textbf{EviRef (Ours)} & full & \underline{96.8} & 94.4 & 87.1 & \underline{88.1} & 84.6 & 90.2 \\
\bottomrule
\end{tabular}
}
\end{table}

Table~\ref{tab:fewnormal} and Fig.~\ref{fig:analysis}(c) through (f) vary the number $n=|\mathcal{N}|$ of normal images used for the memories and standardization statistics. The component prototypes and hyperparameters are those of the full model, and for $n\le10$ the results are averaged over three random subsets. With 10 images per task, EviRef attains 85.8\%, above the method of Nakata \emph{et al.}\ using 50 images (83.1\%) under every condition. With 20 images it reaches 91.5\%, 99\% of the full-data result, and exceeds all compared methods that use 50 normal images per task. The largest gain occurs between 10 and 20 images (5.7 points), whereas the remaining 30 images add only 0.6 points. Already with 20 images, the OCR gate opens on the same five \emph{Ropes} tasks as with all 50. The performance thus saturates quickly, consistent with the introduction's argument that a few dozen images are insufficient to train a discriminator but sufficient to estimate a few low-dimensional statistics reliably.

Fig.~\ref{fig:analysis}(d) and (e) show the effect of the number of samples per path and per capture condition. All three paths stabilize at about 20 images, and R already approaches its full-data performance with 5 images. With few samples, the conditions differ markedly, ranging from 76.5\% to 84.8\% at 5 images, whereas at 20 images all five exceed 90\%, indicating that stable cross-condition performance emerges as the statistics become more reliable. Fig.~\ref{fig:analysis}(f) shows that different constraints differ in their sample requirements. The discrete quantity constraints reach 93.3\% with only 5 images, on a par with the full data, whereas the placement and relation constraints, which involve continuous geometry, need more samples to cover the natural variation of normal layouts. The placement constraints, for example, rise from 69.1\% at 5 images to 98.9\% at 20.

\subsection{Validation on MVTec LOCO AD}
\label{sec:exp_loco}
MVTec LOCO AD has no separate reference condition. Accordingly, the training images of each category serve as both the reference set and the normal set. Instead of the location-wise prototype matching used on VID-AD, component instances are obtained with the same frozen Grounding DINO and SAM-HQ and labeled at test time by nearest-neighbor voting over the training components with WRN-50 features. The memories, standardization, and hyperparameters are those of EviRef, with paths O, V, and R only. Table~\ref{tab:loco_logical} compares training-free methods~\cite{cohen2020spade,roth2022patchcore,liu2023comad,cohen2023sinbad,zhang2025logsad,peng2025samlad,jin2025logicad,kwon2025logicqa}. EviRef attains 90.2\%, below SAM-LAD (95.8\%) and SINBAD (91.2\%) and above the other listed methods. LogicAD and LogicQA use only 1 to 3 reference images but rely on large vision-language models. Per category, EviRef ranks second on \emph{Breakfast Box} (96.8\%) and \emph{Screw Bag} (88.1\%) and has its lowest scores on \emph{Splicing Connectors} (84.6\%) and \emph{Pushpins} (87.1\%).

\section{Conclusion}
\label{sec:conclusion}
We have modeled logical normality as decomposable constraints and realized this view in the training-free EviRef, which shares component prototypes across capture conditions and verifies each evidence family against its own memory estimated from normal images. The experiments support this view in three respects. First, the decomposed constraints reflect the structure of logical anomalies. Each anomaly type is primarily identified by the constraint it violates by definition, with the other constraints providing complementary evidence. Second, EviRef attains the highest mean AUROC among the compared methods under each capture condition of VID-AD, with a standard deviation of 0.5, and its constraints are established from about 20 normal images per task. Third, textual evidence is admitted as a new path without retraining or changing the existing memories, and the same constraint families remain competitive on MVTec LOCO AD. One limitation is that the occupancy and relation descriptors assume a fixed viewpoint and may be sensitive to changes in viewpoint, orientation, or scale. Future work will investigate viewpoint-invariant descriptors.

\bibliographystyle{unsrt}
\bibliography{bib/refs}

\end{document}